\documentclass[conference]{IEEEtran}
\IEEEoverridecommandlockouts

\usepackage{cite}
\usepackage{amsmath,amssymb,amsfonts}
\usepackage{algorithmic}
\usepackage{graphicx}
\usepackage{textcomp}
\usepackage{xcolor}
\usepackage{subcaption}

\usepackage[ruled,vlined,linesnumbered]{algorithm2e}
\usepackage{amsmath}
\usepackage{amsthm}
\usepackage{array}
\usepackage{booktabs, makecell}
\usepackage{calc}
\usepackage{colortbl}
\usepackage{footnote}
\usepackage{graphicx}
\usepackage{hyperref}
\usepackage{cleveref}
\usepackage{listings}
\usepackage{mathtools}
\usepackage{microtype}
\usepackage{multirow}
\usepackage{newfloat}
\usepackage{seqsplit}
\usepackage{subcaption}  
\usepackage{url}
\usepackage{xcolor}
\usepackage{xspace}
\usepackage{lineno}
\usepackage{pifont}
\usepackage{wrapfig}
\usepackage{cleveref}
\usepackage{lipsum}
\usepackage{tcolorbox}
\usepackage{bbold} 

\usepackage{bbm} 

\usepackage[inline,shortlabels]{enumitem} 
\usepackage{siunitx} 
\newcommand{\eg}{{\em e.g.,~}} 
 
\newcommand{\ie}{{\em i.e.,~}}

\newcommand{\myalg}{\texttt{PRSOBG}}

\begin{document}


\title{Physics-Refined Spatiotemporal Forecasting on Open-Boundary Hydrologic Graphs
}

\author{
\IEEEauthorblockN{
Haoyang Jiang\textsuperscript{1},
Zhengui Wang\textsuperscript{2},
Shenghan Gao\textsuperscript{2},
Y. Joseph Zhang\textsuperscript{2},
Xingquan Zhu\textsuperscript{3},
Yi He\textsuperscript{1,$\dagger$}
\thanks{\textsuperscript{$\dagger$}Corresponding author: Yi He (yihe@wm.edu).}
}

\IEEEauthorblockA{
\textsuperscript{1} William \& Mary, Williamsburg, VA, USA;
\{hjiang16, yihe\}@wm.edu\\
\textsuperscript{2}Virginia Institute of Marine Science, William \& Mary,
Gloucester Point, VA, USA;
\{wangzg, sgao, yjzhang\}@vims.edu\\
\textsuperscript{3}Florida Atlantic University, Boca Raton, FL, USA;
xzhu3@fau.edu
}
}

\maketitle

\begin{abstract}
Spatiotemporal forecasting on hydrologic graphs is especially prone to instability 
in \emph{open-boundary} systems, 
where the forecast domain exchanges fluxes with an unobserved exterior.
In such systems, boundary nodes receive external forcing, \eg upstream inflows in rivers or tidal signals in coastal regions, that is typically unavailable at prediction time.
The absence of this information can compound errors 
as forecasts unfold in an autoregressive fashion, leading to inferior 
long-horizon performance.
This paper dissects this instability issue by exploring two questions.
{\em 1) {\bfseries What} boundary forcing enters the forecast domain when
information beyond boundary is missing?}
{\em 2) {\bfseries How} should this forcing propagate through the domain without incurring error amplification under autoregressive rollout?}

To address both, we propose \myalg, a new computing framework 
comprising two key components.
First, to compensate for the 
boundary forcing, \myalg\ learns ghost node proxies
from the boundary and interior nodes,
striving to approximate unobserved external inputs.
Second, to control error accumulation from these learned proxies,
we leverage two physics refiners.
In particular, one refiner enforces {\em local consistency} by aligning ghost proxies with their two-hop neighbors (\ie boundary nodes and their immediate interiors).
The other refiner enhances {\em global stability} by correcting the model forecasts through a physics-guided graph neural operator, reducing long-horizon numerical drift.
Two real-world hydrologic graphs are employed for empirical evaluation.
Comparative results 
show that our proposal enjoys higher prediction accuracy and long-horizon stability
over both learning-based and physics-informed model competitors.
Our code is available at \url{https://github.com/HaoyangJiang-WM/OpenBoundary}.

\end{abstract}


\section{Introduction}

Open-boundary graphs abound in hydrologic modeling and scientific computing, where the forecast domain is only a subregion of larger riverine or coastal systems~\cite{zhang2016schism,klingler2021lamah}.
%
External forcing in such graphs, such as upstream inflows or tidal fluctuations,
enters through boundary nodes but lies outside the modeled region and is unobserved at forecast time.
As a result, standard spatiotemporal models tend to fail in multi-step (autoregressive) prediction,
where even tiny boundary inconsistencies can cascade and compound into substantial downstream 
errors~\cite{bentivoglio2025boundary,mccabe2023stability}.

\begin{figure}[t]
    \centering
    \begin{subfigure}[t]{0.48\columnwidth}
        \centering
        \includegraphics[width=\linewidth]{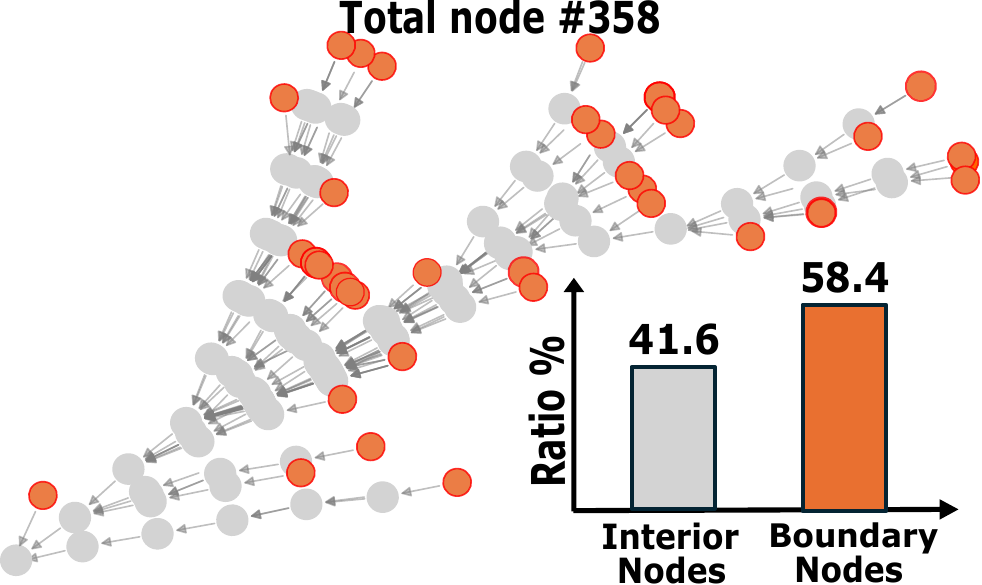}
        \caption{Danube river system with 358 gauge stations in total; boundary nodes (highlighted) comprise 58.4\% of all nodes.}
        \label{fig:river_topo}
    \end{subfigure}\hfill
    \begin{subfigure}[t]{0.48\columnwidth}
        \centering
        \includegraphics[width=\linewidth]{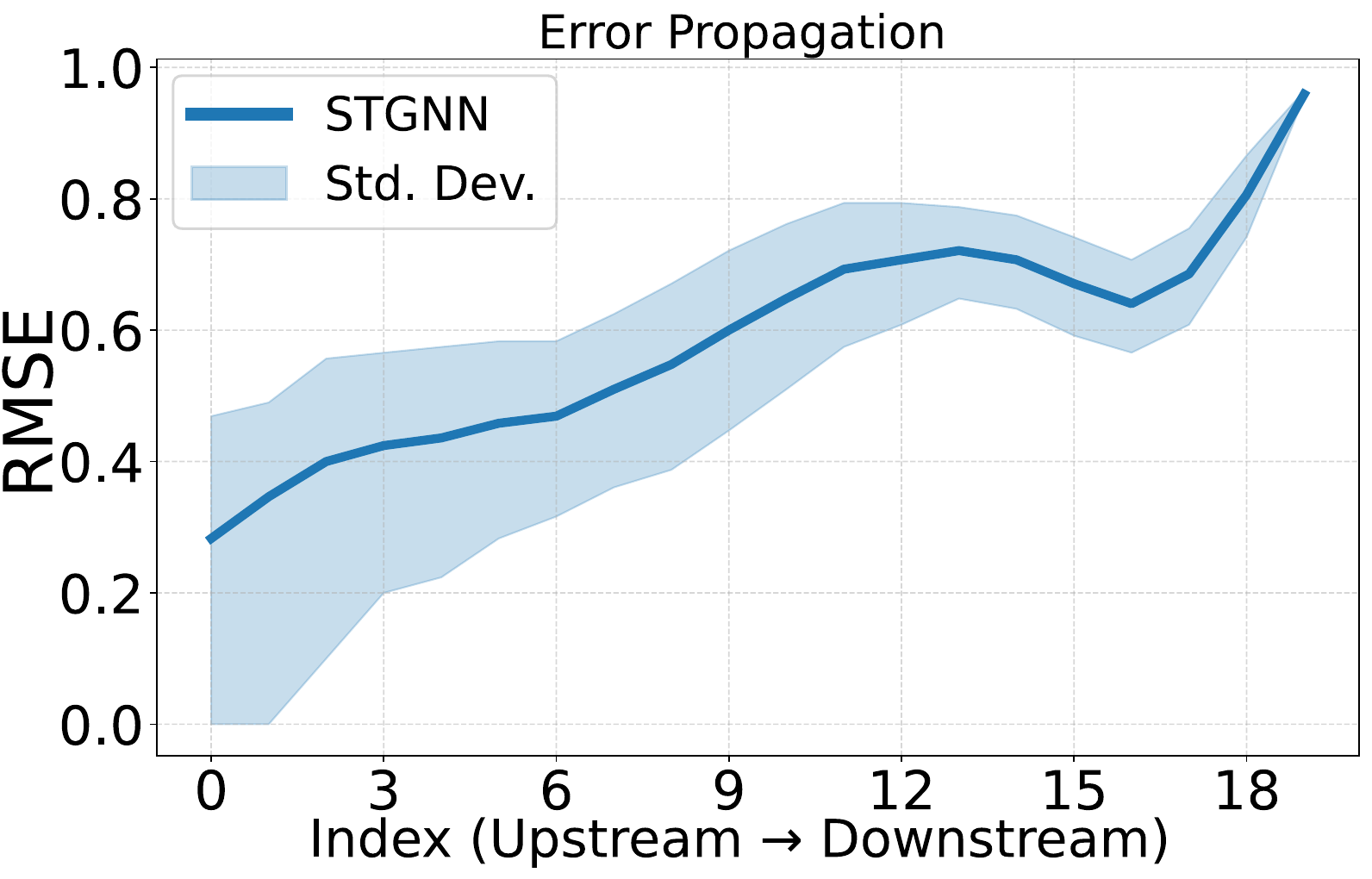}
        \caption{Node-wise RMSE increases; ``0'' indices boundary nodes, and larger index represents farther downstream node.}
        \label{fig:error_prop}
    \end{subfigure}
    \caption{Demonstration of downstream error amplification in the Danube river system~\cite{klingler2021lamah}.
    (a) Boundary nodes regulate the influx of entire network due to unobserved external forcing, and (b) boundary errors propagate and amplify toward downstream under autoregressive rollout in STGNN.} 
    \label{fig:river_topo_errorprop}
    \vspace{-6pt}
\end{figure}

To wit, we train a vanilla spatiotemporal graph neural network (STGNN)~\cite{li2018dcrnn,wu2019graphwavenet}
on the Danube river system in Central Europe~\cite{klingler2021lamah}, 
as shown in Figure~\ref{fig:river_topo_errorprop}(a).
Each node corresponds to a gauge station that monitors  multivariate water variables 
(\eg discharge, flow velocity, and temperature), 
and the task is to forecast future fluxes across the network,
to support flood forecasting and early warning.
As entry points for unobserved external forcing,
the boundary nodes (with no further upstream gauges) are highlighted,
which constitute 58.4\% of the network.
As shown in Figure~\ref{fig:river_topo_errorprop}(b), prediction errors systematically increase from upstream to downstream, as initial errors at boundary nodes propagate and amplify into the interior.
%
This highlights boundary-induced error amplification as a key source of long-horizon degradation.


In this paper, we explore and dissect this forecast instability in open-boundary hydrologic graphs 
through  \underline{two main questions}.
First, {\em \bfseries what} boundary forcing enters the forecast domain when information beyond boundary nodes is in general unobserved?
Second, {\em \bfseries how} should this forcing propagate through the network without inducing error amplification under autoregressive rollout?

To address these challenges, we propose a new framework,
termed \emph{Physics-Refined Spatiotemporal forecasting on Open-Boundary
hydrologic Graphs} (\myalg),
which comprises two key building blocks.
To address the {\em \bfseries what} problem, \myalg\ learns to supplement missing boundary forcing through {\em ghost-node proxies}.
Our key idea is to make the unobserved boundary forcing explicit using auxiliary nodes~\cite{BergerLeVeque98}, 
placed outside the forecast domain,
whose states encode boundary conditions.
Each ghost node aims to infer latent boundary forcing
from observable in-domain dynamics between boundary nodes and 
its immediate interior  (downstream) neighbors.

Yet, such learned ghost-based boundary proxies remain approximate at local regions.
In autoregressive settings, small approximation errors 
can propagate inward and accumulate over time.
%
%
This motivates the complementary {\em \bfseries how} problem.
To this end, we leverage {\em physics refiners}~\cite{lippe2023pde,Huang2025PhysicsCorrect}
to regularize the propagation dynamics in the graph.
At each rollout step, the predicted results are fed into a refinement layer 
before the next iteration.
This refinement proceeds in two stages.
%
Namely, {\bf 1)} a \emph{local} boundary refiner acts on boundary nodes and their neighbors
to suppress a self-reinforcing ghost--boundary error loop:
errors in ghost nodes can contaminate boundary states, which in turn bias
subsequent ghost-node approximations and amplify the error.
The boundary refiner controls this feedback and stabilizes boundary forcing
estimates before propagation.
{\bf 2)} A \emph{global} physics refiner
performs a physics-guided discrete correction over the entire graph.
Rather than solving a full PDE, it applies a structured numerical update based on
the dominant propagation dynamics of the system.
This helps reduce drift and error accumulation under long autoregressive rollout.
We instantiate this global physics refiner in two forms.
First, we adapt an explicit refiner from~\cite{jiang2025topology},
which applies a PDE-inspired graph update to align predictions with dominant propagation dynamics.
Second, we derive a convexified variant that renders the refiner semi-implicit.
It anchors the correction to the boundary-refined prediction at each rollout step,
lending numerical robustness~\cite{leveque2002finite}
when simplified dynamics priors mismatch real hydrologic dynamics.

\noindent\textbf{Specific contributions} in this paper are as follows.

\begin{enumerate}
\item \textbf{Problem Analysis.}
To the best of our knowledge, this is the first study formalizing and decomposing the spatiotemporal 
forecasting on open-boundary graphs as a two-tier {\em what-how} challenge.
%
%
This analysis identifies unresolved boundary forcing as a major source of autoregressive rollout degradation, particularly in hydrologic applications.

\item \textbf{Computing Framework.}
We propose \myalg\ to operationalize the proposed \emph{what-how} decomposition.
\myalg\ augments a standard STGNN backbone 
with ghost-node proxies and physics-guided refiners
to explicitly infer unobserved boundary forcing,
refine this approximation and stabilize its propagation under long-horizon rollout.
%


\item \textbf{Mesoscale Real-World Evaluation.}
We validate \myalg\ on two real hydrologic systems at mesoscale,
the Danube river system (directed graph) 
and the Chesapeake Bay (unstructured mesh).
This testbed targets an unexplored regime between simulation-driven physics-informed models (microscale) and coarse Earth system forecasting (macro), where 
unobserved boundary deficits and irregular graph 
topology jointly exacerbate long-horizon autoregressive instability.
Empirical results demonstrate that \myalg\ outperforms \textbf{9} baselines,
including \textbf{6} spatiotemporal graph learning and \textbf{3} physics-informed competitors,
with an average RMSE reduction of \textbf{38.3\%} over the STGNN backbone across all reported horizons.

\end{enumerate}

\section{Preliminaries}
\label{sec:prelim}

\noindent\textbf{Notation.}
Let $G=(V,E)$ denote the hydrologic graph, where $V=\{v_1,\dots,v_N\}$ is a set of $N$ nodes representing local observation locations and $E\subseteq V\times V$ is the edge set.
At time step $t$, each node $v_i$ has a feature vector $\mathbf{h}_i^{(t)}\in\mathbb{R}^{C}$ with $C$ channels, and we write the node-feature matrix as
$\mathbf{H}^{(t)}=[\mathbf{h}_1^{(t)},\dots,\mathbf{h}_N^{(t)}]^\top\in\mathbb{R}^{N\times C}$.
Each edge $(v_i,v_j)\in E$ is associated with a static edge-feature vector $\mathbf{e}_{ij}\in\mathbb{R}^{D_e}$ with $D_e$ channels encoding physical attributes such as distance.
Connectivity is described by an adjacency matrix $\mathbf{A}\in\{0,1\}^{N\times N}$.

\noindent\textbf{Topological regimes.}
For generality, we consider two topological regimes commonly encountered in real-world hydrologic systems.
(i) Directed graphs, whose edges indicate a prescribed upstream--downstream flow direction, yielding an asymmetric adjacency \(\mathbf A\neq\mathbf A^\top\).
This naturally models river systems, where mass and information propagate directionally according to physical flow constraints, such as the Danube river system~\cite{klingler2021lamah}.
(ii) Unstructured meshes, whose edges represent undirected neighborhood proximity without an intrinsic flow orientation, resulting in a symmetric adjacency \(\mathbf A=\mathbf A^\top\).
This regime is typical of coastal and estuarine hydrodynamic models, where spatial interactions are defined over irregular meshes and directional flow must be inferred from state evolution, such as Chesapeake Bay~\cite{zhang2016schism}.


\noindent\textbf{Open boundary nodes.}
We define open-boundary nodes \(V_{\mathrm{B}}\subseteq V\) by topology.
For directed graphs, \(V_{\mathrm{B}}\) is the set of source nodes with zero in-domain in-degree, i.e., \(\deg^{-}(v_i)=0\).
For unstructured meshes, $V_{\mathrm{B}}$ is pre-labeled by the mesh/model specification (e.g., ocean-connected boundary arcs) and treated as given, indicating where \emph{external forcing} enters the domain.
We denote a boundary node by \(v_b\in V_{\mathrm{B}}\), and define the interior node set as \(V_{\mathrm{I}}:=V\setminus V_{\mathrm{B}}\).


\noindent\textbf{Problem statement.}
Given a lookback window of length $W$, we denote the historical input tensor as
$
\mathcal{H}=\{\mathbf{H}^{(t-W+1)},\dots,\mathbf{H}^{(t)}\}\in\mathbb{R}^{N\times W\times C}.
$
The goal is to predict the target physical quantity over a future horizon of length $T$,
namely 
$
\mathbf{Y}^{(t+1:t+T)}\in\mathbb{R}^{N\times T\times Q},
$
where $Q$ is the number of target channels. 
In our setting, $Q=1$,
and the predictive target is discharge  and Total Water Level (TWL)~\cite{kirschstein2024merit,cui2024total} on directed graph 
and unstructured mesh, respectively. 
Forecasts proceed autoregressively.
At each step the model $f_\theta$ predicts the next state
$\hat{\mathbf{Y}}^{(t+1)}$, and this prediction is recursively fed back as part of the input to generate multi-step forecasts up to $t+T$.
In addition to the historical inputs $\mathcal{H}$, the model $f_\theta$ takes the graph structure  $\mathbf{A}$ and edge attributes $\{\mathbf{e}_{ij}\}_{(i,j)\in E}$ as inputs.
Learning objective is to minimize the rollout loss over the training dataset $\mathcal{D}$:
$$
\min_{\theta}\;
\mathbb{E}_{(\mathcal{H},\,\mathbf{Y}_{GT})\sim \mathcal{D}}
\Big[
\ell\!\left(
\mathbf{Y}_{GT}^{(t+1:t+T)},\;
f_\theta\!\big(\mathcal{H},\,\mathbf{A},\,\{\mathbf{e}_{ij}\}\big)
\right)
\Big],
$$
where $\ell(\cdot)$ denotes the loss metric 
(\eg RMSE).

\begin{figure*}[h]         
  \centering                 
  \includegraphics[width=0.90\textwidth]{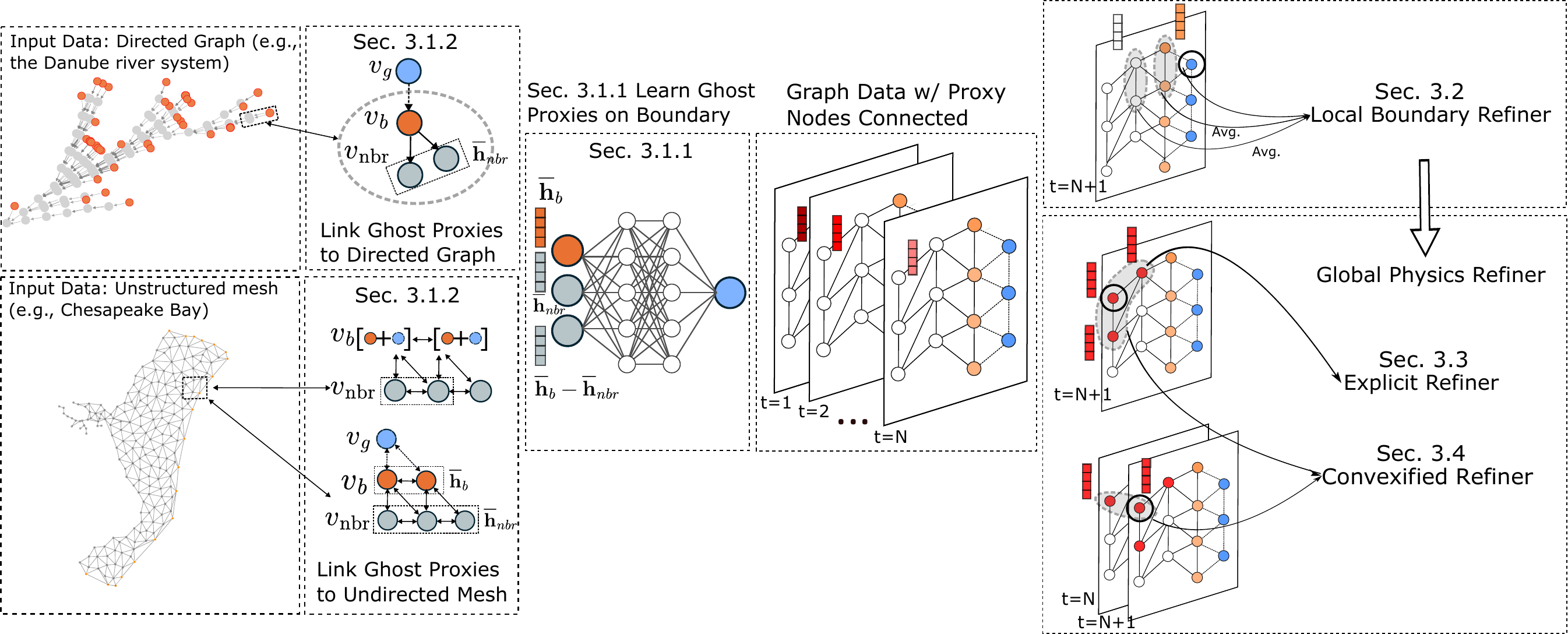} 
\caption{\textbf{Overall pipeline of \myalg.}
For \emph{directed graphs} (e.g., the Danube river system) and \emph{unstructured meshes} (e.g., Chesapeake Bay),
\myalg\ makes missing open-boundary forcing learnable by constructing ghost-node proxies from boundary--interior dynamics (Sec.~\ref{sec:shared_proxy_en_mlg}).
The proxy is instantiated as one-to-one \emph{upstream explicit ghost nodes} in directed graphs, and as three \emph{mesh-specific ghost variants} in unstructured meshes to inject boundary forcing into message passing (Sec.~\ref{sec:three_variants}).
During autoregressive rollout, \myalg\ first applies a local \emph{boundary refiner} to stabilize proxy--boundary consistency (Sec.~\ref{sec:refiner_boundary}), then applies a global \emph{physics refiner} to regularize propagation, using either an \emph{explicit} refiner (Sec.~\ref{sec:refine2}) or a \emph{convexified semi-implicit} refiner (Sec.~\ref{sec:refine3}).}

  \label{fig:model}   
  \vspace{-5pt}
\end{figure*}

\section{Proposed Framework}
\label{sec:method}

We build on a standard STGNN backbone and propose \myalg\ for stable spatiotemporal forecasting on open-boundary hydrologic graphs.
Open-boundary instability is decomposed into two coupled questions:
\emph{what} missing boundary forcing to represent when the exterior is unobserved, and
\emph{how} this forcing should propagate through the forecast domain without amplifying errors under autoregressive rollout.
To address the \emph{what} question, \myalg\ learns \emph{ghost-node proxies} from observable in-domain dynamics between boundary nodes and adjacent in-domain neighbors (Sec.~\ref{sec:open_boundary}), and injects them according to the graph topology (Sec.~\ref{sec:ghost_injection}).
To address the \emph{how} question, \myalg\ uses a \emph{local boundary refiner} to stabilize proxy--boundary consistency (Sec.~\ref{sec:refiner_boundary}) and a \emph{global physics refiner} to correct boundary-driven propagation over the graph (Sec.~\ref{sec:physics_refiner}).
The global physics refiner is instantiated as either an \emph{explicit} or a \emph{convexified semi-implicit refiner}.
An overview is shown in Fig.~\ref{fig:model}.

\vspace{-2pt}
\subsection{\textbf{Learn Boundary Forcing via Proxies}}
\label{sec:open_boundary}

Open-boundary nodes exchange fluxes with an unmodeled exterior, such as upstream inflows in directed river systems and offshore tidal signals in unstructured coastal meshes.
Since these external signals are unavailable at forecast time, standard STGNNs that aggregate only in-domain neighborhoods cannot explicitly represent the missing boundary influence.
To compensate for this missing boundary forcing, we learn ghost-node proxies from observable in-domain dynamics between each boundary node and its adjacent in-domain neighbors, such as immediate downstream neighbors in a directed river system.

Let \(V_{\mathrm{B}}\) be the set of open-boundary nodes and \(V_{\mathrm{I}}=V\setminus V_{\mathrm{B}}\) be the set of interior nodes.
For a boundary node \(b\in V_{\mathrm{B}}\), we define \(V_{\mathrm{nbr}}(b)\subseteq V_{\mathrm{I}}\) as the set of immediate in-domain neighbors adjacent to this boundary node \(b\).
In a directed graph, \(V_{\mathrm{nbr}}(b)\) contains immediate downstream nodes.
In an unstructured mesh, \(V_{\mathrm{nbr}}(b)\) contains one-hop interior nodes near the open boundary.

\textbf{Ghost-Node Boundary Proxy}
\label{sec:shared_proxy_en_mlg}
We motivate the ghost-node boundary proxy from ghost-node discretizations in numerical PDE solvers~\cite{tseng2003ghost}.
We associate each graph node \(v_i\) with a physical location \(x_i\), and write \(u(x_i,t)\) for the state at that location.
For an open boundary location \(x_b\), we represent the unobserved external forcing by a ghost point outside the modeled domain with state \(u_g(t)\).
Consider a Robin-type boundary condition \(\delta_1u(x_b,t)+\delta_2\partial_nu(x_b,t)=u_g(t)\), where \(\partial_n\) is the outward normal derivative.
Using a one-sided approximation with an immediate interior neighbor \(x_{\mathrm{nbr}}\), we have \(\partial_nu(x_b,t)\approx (u(x_b,t)-u(x_{\mathrm{nbr}},t))/\Delta x\), which gives the discrete Robin relation
\begin{equation}
u_g(t)
=
\delta_1u(x_b,t)
+
\delta_2
\frac{u(x_b,t)-u(x_{\mathrm{nbr}},t)}{\Delta x}.
\label{eq:robin_discrete}
\end{equation}
Equation~\eqref{eq:robin_discrete} motivates parameterizing a boundary-forcing surrogate from two observable in-domain quantities: the boundary state and the boundary--interior difference.

In hidden space, for a ghost proxy \(g\), let \(B(g)\subseteq V_{\mathrm{B}}\) be the set of boundary nodes associated with \(g\), and define its corresponding in-domain neighbor set as \(V_{\mathrm{nbr}}(g)=\bigcup_{b\in B(g)}V_{\mathrm{nbr}}(b)\).
At layer \(l\), we summarize the boundary and interior contexts by
\[
\small
\bar{\mathbf h}_{B(g)}^{(l)}
=
\frac{1}{|B(g)|}\sum_{b\in B(g)}\mathbf h_b^{(l)},\;
\bar{\mathbf h}_{\mathrm{nbr}(g)}^{(l)}
=
\frac{1}{|V_{\mathrm{nbr}}(g)|}
\sum_{j\in V_{\mathrm{nbr}}(g)}
\mathbf h_j^{(l)}.
\]
The boundary--interior difference is \(\Delta\mathbf h_g^{(l)}=\bar{\mathbf h}_{B(g)}^{(l)}-\bar{\mathbf h}_{\mathrm{nbr}(g)}^{(l)}\).
The ghost-node boundary proxy is parameterized by a lightweight MLP:
\begin{equation}
\mathbf h_g^{(l)}
=
\Phi_B
\left(
[
\bar{\mathbf h}_{B(g)}^{(l)},
\bar{\mathbf h}_{\mathrm{nbr}(g)}^{(l)},
\Delta\mathbf h_g^{(l)}
];
\theta_B
\right).
\label{eq:ghost_proxy}
\end{equation}
The pair \((\bar{\mathbf h}_{B(g)}^{(l)},\bar{\mathbf h}_{\mathrm{nbr}(g)}^{(l)})\) corresponds to \((u(x_b,t),u(x_{\mathrm{nbr}},t))\) in Eq.~\eqref{eq:robin_discrete}.
Rather than fixing the Robin coefficients in hidden space, \(\Phi_B\) learns an equivalent boundary forcing surrogate.

\subsection{\textbf{Ghost Injection under Different Graph Topologies}}
\label{sec:ghost_injection}
\label{sec:three_variants}

Equation~\eqref{eq:ghost_proxy} gives a unified ghost-node boundary proxy, but its injection should follow the boundary topology.
In directed graphs (e.g., the Danube river system), edges encode a prescribed upstream--downstream flow direction,
so we instantiate a one-to-one upstream ghost node for each open-boundary node.
In unstructured meshes (e.g., the Chesapeake Bay coastal mesh), open-boundary arcs couple to multiple interior neighbors and have no intrinsic flow orientation;
we therefore consider three mesh-specific ghost variants below.

\subsubsection{\textbf{Upstream explicit ghost nodes for directed graphs}}

In directed graphs such as the Danube river system, edges encode a prescribed upstream--downstream flow direction.
For each open-boundary node \(b\), we introduce a one-to-one upstream explicit ghost node \(g(b)\in V_{g}\) and add a directed edge \(g(b)\rightarrow b\) to inject the learned external forcing.
The augmented graph is \(G'=(V\cup V_g,\;E\cup E_{GB})\), where \(E_{GB}=\{(g(b),b)\}\).
At each layer, ghost representations are computed by Eq.~\eqref{eq:ghost_proxy}, and message passing is performed on the augmented graph so that boundary forcing enters along the physical flow direction.
After \(L\) layers, prediction and supervision are applied only to the original nodes: \(\mathbf H_{\mathrm{out}}=\Pi_V(\mathbf H'^{(L)})\), where \(\Pi_V\) discards ghost nodes.
Thus, ghost nodes serve as upstream carriers of boundary forcing and are not directly supervised.

\subsubsection{\textbf{Mesh-specific ghost variants for unstructured meshes}}

Unlike directed graphs, unstructured meshes such as Chesapeake Bay do not provide a prescribed upstream--downstream flow direction.
Their open-boundary arcs are geometrically defined and may couple to multiple interior neighbors, so boundary forcing cannot be injected through a single upstream direction.
Therefore, we provide three mesh-specific ghost variants to connect the learned boundary proxy to boundary neighborhoods and inject it into message passing.
All variants share the same ghost-node boundary proxy in Eq.~\eqref{eq:ghost_proxy}, but differ in how the proxy is injected into the mesh: 
as a residual update to boundary states, as explicit mirrored ghost nodes outside the boundary, or as mirrored ghost nodes with learnable boundary--ghost edges.

\paragraph{\textbf{Variant 1 -- implicit ghost fusion}}
The lowest-cost variant does not add explicit nodes.
Instead, the ghost proxy is injected as a boundary residual:
$
\mathbf h_b^{(l)}
\leftarrow
\mathbf h_b^{(l)}
+
\mathbf h_{g(b)}^{(l)}$,
where $b\in V_{\mathrm{B}}$.
This variant treats the ghost proxy as a boundary residual adapter.

\paragraph{\textbf{Variant 2 -- explicit mirrored ghost layer}}
The second variant explicitly constructs a mirrored ghost layer outside the open boundary.
Let \(\mathbf x_i\in\mathbb R^d\) denote the physical position vector of node \(v_i\).
For each boundary node \(b\), let \(\mathbf x_b\) be its position and \(\mathbf n_b\) be the local outward normal estimated from boundary geometry.
We define a local mirror operator \(\mathcal M_b\) around \(b\) as
\[
\mathcal M_b(\mathbf x)
=
\mathbf x
-
2
\left(
(\mathbf x-\mathbf x_b)^\top \mathbf n_b
\right)
\mathbf n_b .
\]
For each interior neighbor \(j\in V_{\mathrm{nbr}}(b)\), we create a ghost node \(g_b(j)\) with coordinate \(\mathbf x_{g_b(j)}=\mathcal M_b(\mathbf x_j)\).
We mirror each boundary--interior edge into a boundary--ghost edge:
\[
E_{BG}^{\mathrm{mir}}
=
\{
(b\leftrightarrow g_b(j)):
b\in V_{\mathrm{B}},\;
j\in V_{\mathrm{nbr}}(b)
\}.
\]
To preserve local outer geometry, we also mirror the induced interior connectivity:
\[
E_{GG}^{\mathrm{mir}}
=
\{
(g_b(j)\leftrightarrow g_b(j')):
(j\leftrightarrow j')\in E,\;
j,j'\in V_{\mathrm{nbr}}(b)
\}.
\]
The augmented mesh is \(G'_{\mathrm{mir}}=(V\cup V_g,\;E\cup E_{BG}^{\mathrm{mir}}\cup E_{GG}^{\mathrm{mir}})\).
Message passing on \(G'_{\mathrm{mir}}\) allows the learned boundary condition field to interact explicitly with the original mesh.
\paragraph{\textbf{Variant 3 -- mirrored ghost layer with learnable coupling}}
The third variant keeps the mirrored ghost scaffold but makes boundary--ghost coupling learnable.
For each boundary node \(b\), we select a candidate ghost set \(\mathcal C(b)\) by spatial \(k\)NN on the ghost layer.
For each candidate pair \((b,g)\), we define geometric features \(f_{bg}=[\mathbf x_g-\mathbf x_b,\|\mathbf x_g-\mathbf x_b\|_2]\).
Bidirectional gates are learned as
\[
\gamma_{b\to g}
=
\sigma(\rho_{B\to G}(f_{bg})),
\qquad
\gamma_{g\to b}
=
\sigma(\rho_{G\to B}(f_{bg})).
\]
The distance-decayed edge weights are
\[
w_{b\to g}
=
\frac{\gamma_{b\to g}}
{\|\mathbf x_g-\mathbf x_b\|_2+\varepsilon},
\qquad
w_{g\to b}
=
\frac{\gamma_{g\to b}}
{\|\mathbf x_g-\mathbf x_b\|_2+\varepsilon}.
\]
This yields a learnable boundary--ghost edge set \(E_{BG}(\theta)\) and the augmented graph \(G'_{\theta}=(V\cup V_g,\;E\cup E_{GG}^{\mathrm{mir}}\cup E_{BG}(\theta))\).
This variant is more flexible than fixed mirroring while retaining a stable geometric scaffold.

\subsection{\textbf{Boundary Refiner for Proxy Stabilization}}
\label{sec:refiner_boundary}

Ghost-node proxies approximate unobserved open-boundary forcing, but their errors may drift and accumulate during autoregressive rollout.
At each rollout step, the STGNN backbone with proxy injection produces hidden-state estimates \((\hat{\mathbf h}_g,\hat{\mathbf h}_b,\hat{\mathbf h}_n)\); if used directly, these estimates may contaminate subsequent proxy construction and amplify boundary-forcing errors.
We therefore apply a \emph{local boundary refiner} as a differentiable plug-in layer before the next rollout step, solving a small \emph{strictly convex} problem with a \emph{closed-form} ghost update.

From the discrete Robin relation in Eq.~\eqref{eq:robin_discrete}, the ghost--boundary--interior relation can be written as
$
u_g
=
\left(
\delta_1+\frac{\delta_2}{\Delta x}
\right)u_b
-
\frac{\delta_2}{\Delta x}u_n
=
\alpha u_b+\beta u_n,
$
where \(\alpha=\delta_1+\delta_2/\Delta x\) and \(\beta=-\delta_2/\Delta x\).
Here, \(u_n\) denotes the immediate interior state.
In hidden space, \(\mathbf h_n\) denotes the corresponding interior-neighbor representation.
When a ghost proxy is associated with multiple boundary or interior nodes, we use \(\mathbf h_b=\bar{\mathbf h}_{B(g)}\) and \(\mathbf h_n=\bar{\mathbf h}_{\mathrm{nbr}(g)}\).
The three-point residual is \(r(\mathbf h_g,\mathbf h_b,\mathbf h_n)=\mathbf h_g-\alpha\mathbf h_b-\beta\mathbf h_n\).

Given backbone estimates \((\hat{\mathbf h}_g,\hat{\mathbf h}_b,\hat{\mathbf h}_n)\), the local boundary refiner balances three-point consistency with closeness to the backbone estimates:
\[
\begin{aligned}
\min_{\mathbf h_g,\mathbf h_b,\mathbf h_n}\quad
&
\frac{\lambda_R}{2}
\|
\mathbf h_g-\alpha\mathbf h_b-\beta\mathbf h_n
\|_2^2
+
\frac{\lambda_g}{2}
\|
\mathbf h_g-\hat{\mathbf h}_g
\|_2^2 \\
&+
\frac{\lambda_b}{2}
\|
\mathbf h_b-\hat{\mathbf h}_b
\|_2^2
+
\frac{\lambda_n}{2}
\|
\mathbf h_n-\hat{\mathbf h}_n
\|_2^2 ,
\end{aligned}
\]
where \(\lambda_R,\lambda_g,\lambda_b,\lambda_n>0\).
The first term penalizes violation of the local ghost--boundary--interior constraint, while the remaining terms keep the refined variables close to the STGNN estimates.
Since the objective is strictly convex, it admits a unique minimizer.
We use the closed-form ghost component:
$
\tilde{\mathbf h}_g
=
\hat{\mathbf h}_g
-
\frac{\kappa}{\lambda_g}
r(
\hat{\mathbf h}_g,
\hat{\mathbf h}_b,
\hat{\mathbf h}_n
),
$
where \(\kappa=(\lambda_R^{-1}+\lambda_g^{-1}+\alpha^2\lambda_b^{-1}+\beta^2\lambda_n^{-1})^{-1}\).
This update moves the ghost proxy toward local boundary consistency without overwriting the backbone estimate, and can be inserted as a single closed-form correction layer in each rollout step.

\subsection{\textbf{Global Physics Refiner for Boundary-Driven Propagation}}
\label{sec:physics_refiner}

Beyond the local boundary refiner, which stabilizes local proxy--boundary consistency, we introduce a \emph{global physics refiner} to regularize boundary-driven propagation over the graph.
At each autoregressive rollout step, it takes the boundary-refined representation and applies a structured post-update before readout and feedback, reducing drift accumulation without solving the underlying PDE to convergence.


\subsubsection{\textbf{Explicit Refiner}}
\label{sec:refine2}

We first define the explicit form of the topology-specific post-update.

\paragraph{\textbf{Directed graph}}
For directed graphs, we use an advective prior \(\partial_t u+u\partial_xu=-g_0\partial_xz\), where \(u\) is the along-channel state, \(z\) is bed elevation, and \(g_0\) is gravitational acceleration.
The spatial derivatives are discretized by directed upwind graph derivative operators.
Specifically, \(D_1\) mainly encodes along-channel spacing, while \(D_2\) additionally incorporates elevation-related information to capture slope-driven differences.

Let \(\tilde{\mathbf h}\) denote the boundary-refined graph-wide representation after the local boundary refiner.
At each rollout step, the explicit refiner applies the following post-update:
\[
\tilde{\mathbf h}^{\star}
=
\tilde{\mathbf h}
-
\Delta t
\sigma
\left(
\tilde{\mathbf h}\odot(D_1\tilde{\mathbf h}W_1)
+
\widehat g(D_2\tilde{\mathbf h}W_2)
\right),
\]
where \(W_1,W_2\) and \(\widehat g\) are learnable parameters.
This update is a hidden-space graph-operator approximation of the transport and slope-driven terms.
The corrected state \(\tilde{\mathbf h}^{\star}\) is used for readout and autoregressive feedback.

\paragraph{\textbf{Unstructured mesh}}
For unstructured meshes, we use a linearized barotropic-wave prior.
Let \(\eta\) be free-surface perturbation, \(\mathbf u\) the depth-averaged velocity, \(H\) the still-water depth, and \(r\) a linear friction coefficient.
Starting from the linearized depth-averaged shallow-water equations~\cite{vreugdenhil1994numerical,leveque2002finite},
\[
\partial_t\eta+\nabla\cdot(H\mathbf u)\approx0,
\qquad
\partial_t\mathbf u=-g_0\nabla\eta-r\mathbf u+\mathbf F.
\]
Taking a time derivative of the continuity equation and substituting the momentum equation gives
\[
\partial_t^2\eta+r\partial_t\eta-\nabla\cdot(g_0H\nabla\eta)
=
-\nabla\cdot(H\mathbf F).
\]
For slowly varying or locally constant depth, this reduces to a damped wave equation with \(c^2=g_0H\).
Since \(\mathbf F\) is absorbed by the data-driven backbone, we use the homogeneous part
\(\partial_t^2\eta+r\partial_t\eta-c^2\nabla^2\eta=0\)
as a stabilizing propagation prior.

Let \(\tilde{\mathbf h}\) denote the boundary-refined graph-wide representation at the current rollout step, and let \(\tilde{\mathbf h}^{\mathrm{prev}}\) denote the previous rollout representation.
The mesh explicit refiner applies the following post-update:
\[
\tilde{\mathbf h}^{\star}
=
a_0\sigma(A\tilde{\mathbf h}W_0)
+
a_1\sigma(A\tilde{\mathbf h}^{\mathrm{prev}}W_1)
+
\gamma\sigma(L\tilde{\mathbf h}W_\Delta),
\]
where \(A\) is the adjacency matrix, \(L\) is the hidden-space graph Laplacian, \(W_0,W_1,W_\Delta\) are learnable weight matrices, and \(a_0,a_1,\gamma\) control temporal memory and spatial propagation strength.
The corrected state \(\tilde{\mathbf h}^{\star}\) is used for readout and autoregressive feedback.

\subsubsection{\textbf{Convexified Semi-Implicit Refiner}}
\label{sec:refine3}

Although the explicit refiner is lightweight, its simplified dynamics prior can be mismatched in complex open-boundary systems, especially on high-resolution graphs or meshes where local errors are more easily amplified into long-horizon drift.
To improve robustness, we adopt a semi-implicit update and convexify it into a strictly convex quadratic form, where the boundary-refined prediction acts as an anchor and the dominant dynamics prior acts as a stabilizing structural regularizer~\cite{leveque2002finite,ascher1997implicit}.
This yields an SPD linear system that can be efficiently approximated by a differentiable PCG layer.

\paragraph{\textbf{Directed graph}}
For directed graphs, the convexified semi-implicit refiner is applied directly to the boundary-refined hidden representation.
Let \(\tilde{\mathbf h}\) denote the input representation after the local boundary refiner.
Motivated by the advective prior, we use a semi-implicit refinement: the multiplicative state is fixed at \(\tilde{\mathbf h}\), while the directed derivative is evaluated at the updated state \(\tilde{\mathbf h}^{\star}\):
\[
\tilde{\mathbf h}^{\star}
=
\tilde{\mathbf h}
-
\Delta t
\left[
\tilde{\mathbf h}\odot(D_1\tilde{\mathbf h}^{\star})
+
\widehat gD_2\tilde{\mathbf h}
\right].
\]
Here, \(D_1\) is the directed upwind derivative operator for the semi-implicit transport term, and \(D_2\) encodes the slope-related correction on the current boundary-refined state \(\tilde{\mathbf h}\).
Since \(D_1\tilde{\mathbf h}^{\star}\) uses the updated state, the update is semi-implicit.

The semi-implicit update can be written as \(A_r\tilde{\mathbf h}^{\star}=\mathbf b_r\), where
\(A_r=I+\Delta t\,\mathrm{diag}(\tilde{\mathbf h})D_1\) and
\(\mathbf b_r=\tilde{\mathbf h}-\Delta t\,\widehat gD_2\tilde{\mathbf h}\).
Since directed upwind operators are generally non-symmetric, directly solving \(A_r\tilde{\mathbf h}^{\star}=\mathbf b_r\) can be sensitive.
We therefore solve the convexified objective
\[
\tilde{\mathbf h}^{\star}
=
\arg\min_{\mathbf h}
\frac{1}{2}\|A_r\mathbf h-\mathbf b_r\|_2^2
+
\frac{\lambda}{2}\mathbf h^\top L\mathbf h
+
\frac{\mu}{2}\|\mathbf h-\tilde{\mathbf h}\|_2^2 .
\]
Here, \(L\succeq0\) is a graph Laplacian regularizer,
\(\lambda>0\) controls graph regularization, and
\(\mu>0\) is a small damping coefficient that anchors the solution to the boundary-refined prediction \(\tilde{\mathbf h}\).
The optimality condition gives \(K_r\tilde{\mathbf h}^{\star}=\mathbf y_r\), with
\(K_r=A_r^\top A_r+\lambda L+\mu I\) and
\(\mathbf y_r=A_r^\top\mathbf b_r+\mu\tilde{\mathbf h}\).
Because \(\mu>0\), \(K_r\) is strictly positive definite and can be solved stably by PCG.

\paragraph{\textbf{Unstructured mesh}}
For the barotropic-wave prior on an unstructured mesh, let \(L_g=-\Delta_g\succeq0\) denote the physical positive semidefinite graph operator, and let \(L\) denote its hidden-space implementation.
To make the semi-implicit structure explicit, we first write the update as
\[
\boldsymbol\eta^{n+1}
=
\boldsymbol\eta^n
+
\gamma(\boldsymbol\eta^n-\boldsymbol\eta^{n-1})
-
\beta L_g\boldsymbol\eta^{n+1}.
\]
Here, the spatial propagation term \(L_g\boldsymbol\eta^{n+1}\) uses the updated state, so this is an implicit spatial update.
Rearranging gives
\((I+\beta L_g)\boldsymbol\eta^{n+1}
=
\boldsymbol\eta^n+\gamma(\boldsymbol\eta^n-\boldsymbol\eta^{n-1})\).
In hidden space, this becomes \(K_b\tilde{\mathbf h}^{\star}=\mathbf y_b\), where \(K_b=I+\beta L\) and
$
\mathbf y_b
=
a_0\sigma(A\tilde{\mathbf h}W_0)
+
a_1\sigma(A\tilde{\mathbf h}^{\mathrm{prev}}W_1).
$
Equivalently, the mesh refiner minimizes
$
\mathcal E(\mathbf h)
=
\frac{1}{2}\|\mathbf h-\mathbf y_b\|_2^2
+
\frac{\beta}{2}\mathbf h^\top L\mathbf h.
$
Since \(I+\beta L\) is SPD for \(\beta\ge0\), this update is stable and efficiently solvable.

\subsubsection{\textbf{Differentiable PCG Layer}}

Both convexified semi-implicit refiners reduce to a unified SPD system \(K\tilde{\mathbf h}^{\star}=\mathbf y\), with \((K,\mathbf y)=(K_r,\mathbf y_r)\) for directed graphs and \((K_b,\mathbf y_b)\) for unstructured meshes.
We approximate the solution using a fixed number of differentiable PCG iterations.

The warm start is the boundary-refined representation, with
\(\tilde{\mathbf h}^{(0)}=\tilde{\mathbf h}\) and
\(\mathbf r^{(0)}=\mathbf y-K\tilde{\mathbf h}^{(0)}\).
We use a diagonal preconditioner \(M=\mathrm{diag}(\mathbf m)+\varepsilon I\), where \(\mathbf m\) is parameterized to be nonnegative.
After \(K_{\mathrm{pcg}}\) iterations, the approximate solution
\(\tilde{\mathbf h}^{\star}\approx\tilde{\mathbf h}^{(K_{\mathrm{pcg}})}\)
is used as the globally refined hidden state for readout and the next autoregressive step.


\section{Experiments}
\label{sec:experiments}

\noindent\textbf{Datasets.}
We evaluate on two hydrologic forecasting benchmarks with different boundary/topology properties.
\textbf{(i) Directed graph.}
We use a real-world river system preprocessed from LamaH-CE2 over the Danube basin~\cite{klingler2021lamah}, which provides hourly discharge observations together with meteorological records.
The graph contains \(358\) nodes and \(357\) directed edges, and is partitioned into \(209\) open-boundary source nodes and \(149\) interior nodes.
Each node is associated with time-varying discharge.
Each edge is associated with distance, elevation, and slope.
The prediction target is the future discharge at each node.
Forecasting discharge is central to flood forecasting and early warning, enabling timely response and risk reduction~\cite{Kazadi2024FloodGNNGRU,kirschstein2024merit}.
\textbf{(ii) Unstructured mesh.}
We use a high-resolution unstructured mesh of Chesapeake Bay with a \(30\)-minute sampling interval.
The mesh has \(51{,}215\) nodes, \(72{,}877\) faces, and \(124{,}118\) edges.
Open-boundary nodes are the mesh vertices along the ocean boundary, totaling \(191\) nodes.
Each node includes four input channels: total water level (TWL), depth, and a simple time encoding given by \(\sin\) and \(\cos\).
The prediction target is the future TWL at each node.
Forecasting TWL supports coastal flooding and erosion hazard assessment~\cite{cui2024total}.

\medskip
\noindent\textbf{Metrics.}
We evaluate supervised spatiotemporal node regression with autoregressive rollout.
Given a window of length \(W\), the model predicts the next step and recursively feeds predictions back to obtain a \(T\)-step forecast.
We use an input window of \(W{=}24\) and evaluate rollout horizons \(T{=}6\), \(T{=}16\), and \(T{=}30\).
We report node- and rollout-averaged RMSE for Danube discharge prediction and Chesapeake Bay TWL prediction.
For relative comparison, we compute improvement with respect to the STGNN backbone under the same horizon.

\medskip
\noindent\textbf{Competitors.}
We compare against two categories of methods.
\emph{Data-driven baselines} include GRU, a temporal-only baseline; STGNN (Backbone), our spatiotemporal GNN predictor without ghost-node proxies or physics refiners; and Ghost-Only STGNN, which uses ghost-node proxies but disables the boundary refiner and physics refiners.
For Ghost-Only STGNN, we use explicit upstream ghost nodes on the River System and report Ghost Variants~1--3 on Chesapeake Bay.
We further include representative temporal-graph forecasting architectures:
GraphNODE~\cite{Poli2020GraphNeuralODE},
GraphSSM~\cite{Li2024SSMTemporalGraphs},
CTDG~\cite{Gravina2024LongRangeCTDG},
and STPGNN~\cite{Kong2024STPGNN}.
\emph{Physics-informed baselines} include methods that incorporate physical priors or physics-informed correction during autoregressive rollout:
FAIR~\cite{luo2023fair},
PDE-Refiner~\cite{lippe2023pde},
and PhysicsCorrect (PC)~\cite{Huang2025PhysicsCorrect}.
For all competitors, we follow the original papers for model configurations when applicable and report results under the same train/validation/test splits and autoregressive rollout setting.

\medskip
\noindent\textbf{Implementation.}
We use a unified autoregressive protocol with history window \(W\) and rollout horizon \(T\).
All variants of \myalg\  share the same STGNN backbone with hidden size \(d{=}128\); ablations only toggle ghost-node proxies and physics refiners.
All refiner variants are evaluated with ghost-node proxies.
For Chesapeake Bay, we use Ghost Variant~1, namely implicit ghost fusion, as the lightweight default configuration unless otherwise specified.
The ghost-node boundary proxy is produced by a 2-layer MLP \(\Phi_B\).
Prediction and loss are computed only on the original nodes \(V\); ghost nodes, when present, are intermediate carriers and are not directly supervised.
For the Explicit refiner, both the directed-graph and unstructured-mesh settings apply a structured post-update after the backbone output.
For the Convexified refiner, we solve the resulting SPD system using a differentiable unrolled PCG layer with two iterations by default.
All learnable components are trained end-to-end with Adam under the same training budget.

\begin{table*}[t]
\centering
\caption{River System under autoregressive rollout with \(T\in\{6,16,30\}\).
Imp. is relative to STGNN (Backbone).}
\label{tab:river_results_h16}
\small
\setlength{\tabcolsep}{4.2pt}
\renewcommand{\arraystretch}{1.08}
\setlength{\aboverulesep}{0.15ex}
\setlength{\belowrulesep}{0.15ex}
\setlength{\cmidrulesep}{0.15ex}
\begin{tabular}{l cc cc cc}
\toprule
& \multicolumn{2}{c}{\(T=6\)}
& \multicolumn{2}{c}{\(T=16\)}
& \multicolumn{2}{c}{\(T=30\)} \\
\cmidrule(lr){2-3}\cmidrule(lr){4-5}\cmidrule(lr){6-7}
\textbf{Method}
& \textbf{RMSE} & \textbf{Imp.}
& \textbf{RMSE} & \textbf{Imp.}
& \textbf{RMSE} & \textbf{Imp.} \\
\midrule
\multicolumn{7}{l}{\textit{Data-Driven Baselines}} \\
STGNN (Backbone) & 0.3499 & -- & 0.4147 & -- & 0.5924 & -- \\
GRU (Temporal Only) & 0.2623 & +25.04\% & 0.3746 & +9.67\% & 0.5529 & +6.67\% \\
Ghost-Only STGNN & 0.2777 & +20.63\% & 0.3878 & +6.49\% & 0.4836 & +18.37\% \\
GraphNODE~\cite{Poli2020GraphNeuralODE} & 0.3466 & +0.94\% & 0.4104 & +1.04\% & 0.5871 & +0.89\% \\
GraphSSM~\cite{Li2024SSMTemporalGraphs} & 0.3442 & +1.63\% & 0.4092 & +1.33\% & 0.5815 & +1.84\% \\
CTDG~\cite{Gravina2024LongRangeCTDG} & 0.3479 & +0.57\% & 0.4113 & +0.82\% & 0.5859 & +1.10\% \\
STPGNN~\cite{Kong2024STPGNN} & 0.3543 & -1.26\% & 0.4110 & +0.89\% & 0.5898 & +0.44\% \\
\midrule
\multicolumn{7}{l}{\textit{Physics-Informed Baselines}} \\
FAIR~\cite{luo2023fair} & 0.2955 & +15.55\% & 0.3641 & +12.20\% & 0.5237 & +11.60\% \\
PDE-Refiner~\cite{lippe2023pde} & 0.2385 & +31.84\% & 0.3153 & +23.97\% & 0.4549 & +23.21\% \\
\underline{PhysicsCorrect (PC)}~\cite{Huang2025PhysicsCorrect}
& \underline{0.2341} & \underline{+33.10\%}
& \underline{0.3110} & \underline{+25.01\%}
& 0.4462 & +24.68\% \\
\midrule
\multicolumn{7}{l}{\textit{Ours (Proposed Framework)}} \\
Ghost with Boundary Refiner & 0.2548 & +27.18\% & 0.3565 & +14.03\% & 0.4679 & +21.02\% \\
Ours-Explicit
& 0.2390 & +31.69\%
& 0.3145 & +24.16\%
& \underline{0.4386} & \underline{+25.96\%} \\
\textbf{Ours-Convexified}
& \textbf{0.2311} & \textbf{+33.95\%}
& \textbf{0.3079} & \textbf{+25.75\%}
& \textbf{0.4268} & \textbf{+27.95\%} \\
\bottomrule
\end{tabular}
\vspace{-6pt}
\end{table*}

\begin{table*}[t]
\centering
\caption{Chesapeake Bay under autoregressive rollout with \(T\in\{6,16,30\}\).
Imp. is relative to STGNN (Backbone).}
\label{tab:bay_results}
\small
\setlength{\tabcolsep}{4.2pt}
\renewcommand{\arraystretch}{1.08}
\setlength{\aboverulesep}{0.15ex}
\setlength{\belowrulesep}{0.15ex}
\setlength{\cmidrulesep}{0.15ex}
\begin{tabular}{l cc cc cc}
\toprule
& \multicolumn{2}{c}{\(T=6\)}
& \multicolumn{2}{c}{\(T=16\)}
& \multicolumn{2}{c}{\(T=30\)} \\
\cmidrule(lr){2-3}\cmidrule(lr){4-5}\cmidrule(lr){6-7}
\textbf{Method}
& \textbf{RMSE} & \textbf{Imp.}
& \textbf{RMSE} & \textbf{Imp.}
& \textbf{RMSE} & \textbf{Imp.} \\
\midrule
\multicolumn{7}{l}{\textit{Data-Driven Baselines}} \\
STGNN (Backbone) & 0.0732 & -- & 0.1797 & -- & 0.2864 & -- \\
GRU (Temporal Only) & 0.0524 & +28.42\% & 0.1202 & +33.11\% & 0.2108 & +26.40\% \\
GraphNODE~\cite{Poli2020GraphNeuralODE} & 0.0721 & +1.50\% & 0.1743 & +3.01\% & 0.2782 & +2.86\% \\
GraphSSM~\cite{Li2024SSMTemporalGraphs} & 0.0705 & +3.69\% & 0.1716 & +4.51\% & 0.2746 & +4.12\% \\
CTDG~\cite{Gravina2024LongRangeCTDG} & 0.0718 & +1.91\% & 0.1752 & +2.50\% & 0.2801 & +2.20\% \\
STPGNN~\cite{Kong2024STPGNN} & 0.0730 & +0.27\% & 0.1784 & +0.72\% & 0.2847 & +0.59\% \\
\midrule
\multicolumn{7}{l}{\textit{Ghost Variants on Unstructured Mesh}} \\
Ghost Variant 1 (Implicit) & 0.0509 & +30.46\% & 0.1224 & +31.89\% & 0.1981 & +30.83\% \\
Ghost Variant 2 (Mirror) & 0.0506 & +30.87\% & 0.1211 & +32.61\% & 0.1956 & +31.70\% \\
Ghost Variant 3 (Learnable) & 0.0510 & +30.33\% & 0.1227 & +31.72\% & 0.1994 & +30.38\% \\
\midrule
\multicolumn{7}{l}{\textit{Physics-Informed Baselines}} \\
FAIR~\cite{luo2023fair} & 0.0496 & +32.24\% & 0.1129 & +37.17\% & 0.1817 & +36.56\% \\
PDE-Refiner~\cite{lippe2023pde} & 0.0485 & +33.74\% & 0.1086 & +39.57\% & 0.1742 & +39.18\% \\
PhysicsCorrect (PC)~\cite{Huang2025PhysicsCorrect}
& 0.0468 & +36.07\%
& 0.1045 & +41.85\%
& 0.1669 & +41.72\% \\
\midrule
\multicolumn{7}{l}{\textit{Ours (Proposed Framework)}} \\
Ghost with Boundary Refiner & 0.0502 & +31.42\% & 0.1176 & +34.55\% & 0.1885 & +34.18\% \\
Ours-Explicit
& \underline{0.0442} & \underline{+39.62\%}
& \underline{0.0919} & \underline{+48.86\%}
& \underline{0.1447} & \underline{+49.48\%} \\
\textbf{Ours-Convexified}
& \textbf{0.0432} & \textbf{+40.98\%}
& \textbf{0.0897} & \textbf{+50.08\%}
& \textbf{0.1398} & \textbf{+51.19\%} \\
\bottomrule
\end{tabular}
\end{table*}

\begin{figure}[t]
    \centering
    \begin{subfigure}[t]{0.48\columnwidth}
        \centering
        \raisebox{1.2em}{%
            \includegraphics[width=\linewidth]{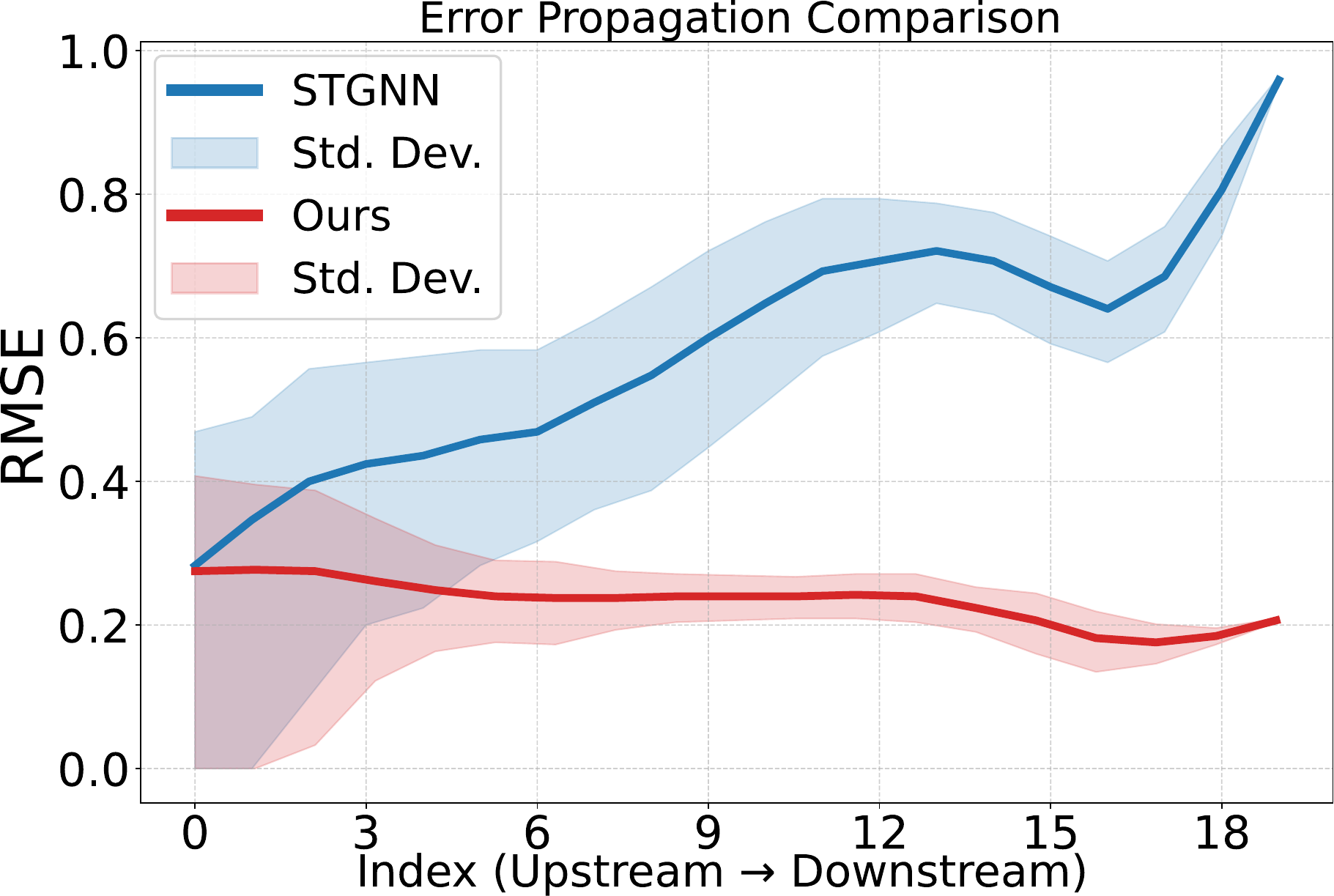}%
        }
        \caption{STGNN exhibits downstream error drift under rollout; \myalg\ keeps errors low.}
        \label{fig:river_rollout_error}
    \end{subfigure}%
    \hspace{0.04\columnwidth}%
    \begin{subfigure}[t]{0.48\columnwidth}
        \centering
        \includegraphics[width=\linewidth]{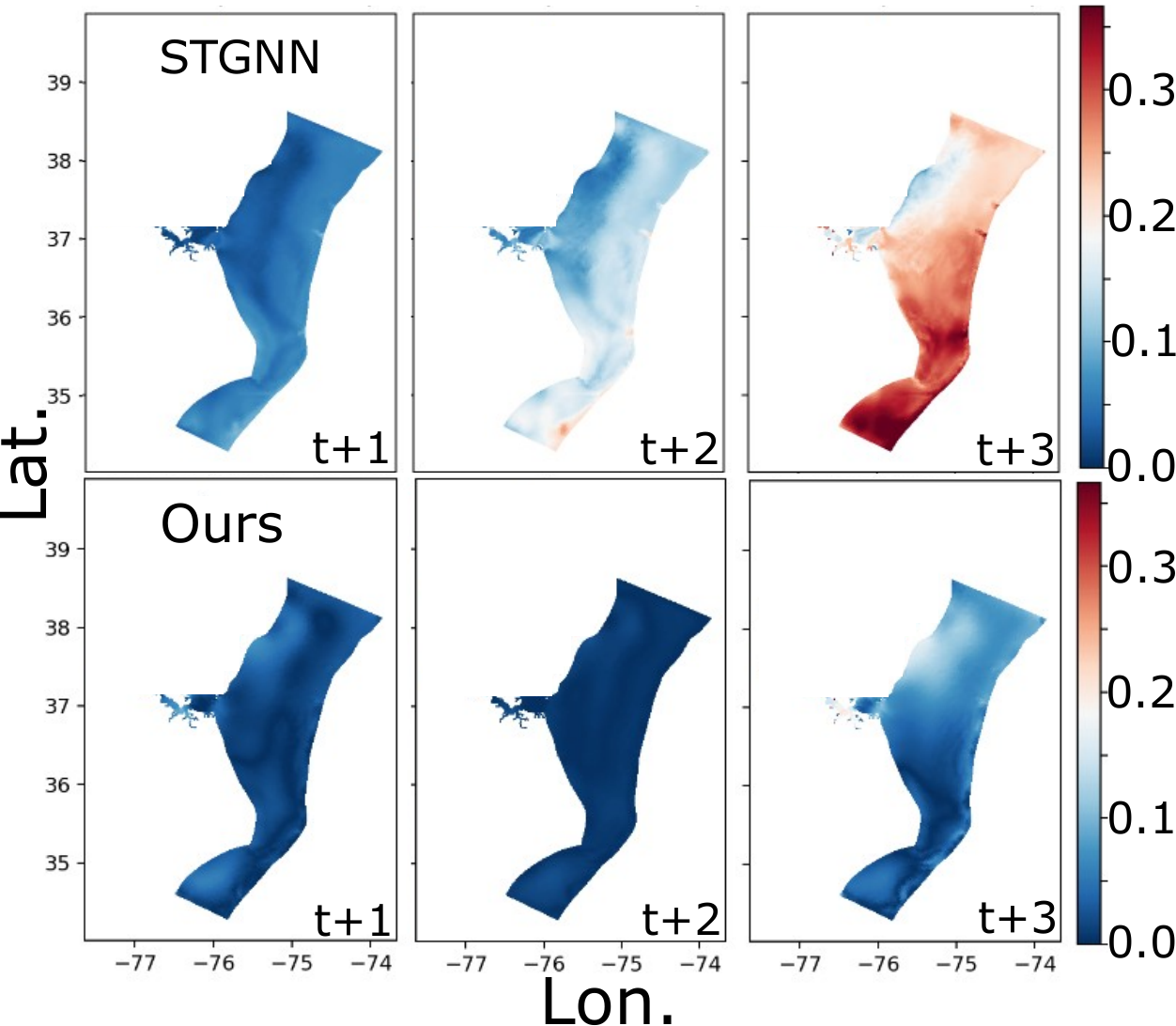}
        \caption{STGNN shows spatial RMSE drift under rollout; \myalg\ suppresses drift.}
        \label{fig:mesh_error_prop}
    \end{subfigure}
    \caption{\textbf{Comparison under autoregressive rollout.}
    Across both directed graphs and unstructured meshes, \myalg\ mitigates boundary-induced instability and achieves consistently lower errors than a vanilla STGNN.}
    \label{fig:rollout_error_compare}
    \vspace{-16pt}
\end{figure}

\medskip
\noindent\textbf{Results and Analysis.}
Based on Table~\ref{tab:river_results_h16}, Table~\ref{tab:bay_results}, and Fig.~\ref{fig:rollout_error_compare}, we answer six research questions.
\emph{RQ1} examines boundary-induced error accumulation.
\emph{RQ2} evaluates ghost proxies and the boundary refiner.
\emph{RQ3} compares explicit and convexified physics refiners.
\emph{RQ4} validates whether the ghost proxy learns a meaningful boundary surrogate.
\emph{RQ5} tests robustness under extreme/high-flow regimes.
\emph{RQ6} studies sensitivity to physical initialization and proxy-network architecture.

\medskip
\noindent\textbf{RQ1. Boundary information deficit and error accumulation.}
On \textbf{\textit{River System}} at \(T{=}6\), the STGNN backbone has RMSE 0.3499, whereas the temporal-only GRU records 0.2623, 25.04\% lower than the backbone.
Since both models use the same GRU temporal backbone, the inferior performance of STGNN indicates that adding spatial message passing alone does not resolve the missing open-boundary information and can amplify boundary-induced errors during rollout.
The pattern persists at \(T{=}16\), where the backbone records 0.4147 and GRU records 0.3746.
At \(T{=}30\), the backbone RMSE further increases to 0.5924, showing increasingly severe error accumulation under longer autoregressive rollout.
Fig.~\ref{fig:rollout_error_compare}(a) further shows that errors increase downstream on the directed river system.
On \textbf{\textit{Chesapeake Bay}}, the same trend holds across all reported horizons.
The backbone records RMSE values of 0.0732, 0.1797, and 0.2864 at \(T{=}6\), \(T{=}16\), and \(T{=}30\), while GRU records 0.0524, 0.1202, and 0.2108, yielding 28.42\%, 33.11\%, and 26.40\% lower errors.
This again shows that adding spatial interactions alone does not resolve the missing open-boundary information.
Fig.~\ref{fig:rollout_error_compare}(b) illustrates that boundary inconsistencies can compound into spatial errors under autoregressive rollout on the unstructured mesh.


\medskip
\noindent\textbf{RQ2. Ghost proxies and boundary refiner.}
Ghost-based boundary proxies directly target the missing-boundary-forcing problem.
On \textbf{\textit{River System}}, Ghost-Only reduces RMSE from 0.3499 to 0.2777 at \(T{=}6\), achieving a 20.63\% improvement over the STGNN backbone.
At \(T{=}16\), it reduces RMSE from 0.4147 to 0.3878.
At \(T{=}30\), it further reduces RMSE from 0.5924 to 0.4836, yielding an 18.37\% improvement.
This indicates that explicitly representing missing source-node boundary forcing helps reduce persistent boundary-side error injection under both short and long rollout.

Adding the boundary refiner further improves stability.
On \textbf{\textit{River System}}, Ghost with Boundary Refiner records RMSE values of 0.2548, 0.3565, and 0.4679 at \(T{=}6\), \(T{=}16\), and \(T{=}30\), improving over Ghost-Only at all reported horizons.
This supports the role of local ghost--boundary--interior consistency in stabilizing the learned boundary proxy before it propagates inward.

On \textbf{\textit{Chesapeake Bay}}, all three mesh-specific ghost variants substantially improve over the STGNN backbone.
At \(T{=}6\), the three variants reduce RMSE from 0.0732 to around 0.051; at \(T{=}16\), they reduce RMSE from 0.1797 to around 0.122; and at \(T{=}30\), they reduce RMSE from 0.2864 to around 0.196--0.199.
With the boundary refiner, RMSE further reaches 0.0502, 0.1176, and 0.1885 at the three horizons.
This demonstrates that the ghost mechanism is not limited to directed river graphs; it also transfers to unstructured meshes where open boundaries are defined geometrically rather than by upstream--downstream topology.
Overall, these results support our \emph{what}-component: explicitly learning a boundary surrogate is important when external forcing is unobserved.

\medskip
\noindent\textbf{RQ3. Explicit and convexified physics refiners.}
Physics refiners address the \emph{how}-component: how the inferred boundary influence should propagate through the graph without amplifying rollout errors.
Across both datasets, Ours-Explicit improves over Ghost-Only and Ghost with Boundary Refiner, showing that an executable physics-guided correction can stabilize propagation beyond local boundary consistency.

On \textbf{\textit{River System}}, Ours-Convexified achieves RMSE values of 0.2311, 0.3079, and 0.4268 at \(T{=}6\), \(T{=}16\), and \(T{=}30\), yielding 33.95\%, 25.75\%, and 27.95\% improvements over the STGNN backbone.
It also improves over the strongest physics-informed baseline, PhysicsCorrect, at all three horizons: 0.2341 vs. 0.2311 at \(T{=}6\), 0.3110 vs. 0.3079 at \(T{=}16\), and 0.4462 vs. 0.4268 at \(T{=}30\).
Ours-Explicit is also competitive, reaching 0.4386 at \(T{=}30\), but the convexified variant remains best.
This shows that the convexified refiner remains effective when autoregressive error accumulation becomes more severe.

On \textbf{\textit{Chesapeake Bay}}, both refiners outperform physics-informed baselines across all horizons, with larger gains under longer rollout.
Ours-Explicit records RMSE values of 0.0442, 0.0919, and 0.1447, while Ours-Convexified further improves them to 0.0432, 0.0897, and 0.1398 at \(T{=}6\), \(T{=}16\), and \(T{=}30\).
These correspond to 40.98\%, 50.08\%, and 51.19\% improvements over the STGNN backbone for Ours-Convexified.
Compared with PhysicsCorrect, Ours-Convexified reduces RMSE from 0.0468 to 0.0432 at \(T{=}6\), from 0.1045 to 0.0897 at \(T{=}16\), and from 0.1669 to 0.1398 at \(T{=}30\).

Overall, the convexified semi-implicit refiner consistently outperforms the explicit refiner.
This suggests that anchoring the correction to the boundary-refined prediction while imposing a stable physics-guided structure is more robust than a purely explicit update under long autoregressive rollout. 

\begin{table}[h]
\centering
\caption{Computational cost relative to the STGNN.}
\label{tab:runtime}
\small
\setlength{\tabcolsep}{2.6pt}
\renewcommand{\arraystretch}{1.03}
\begin{tabular}{lccc}
\toprule
\textbf{Method}
& \textbf{Train (rel.)}
& \textbf{Inference (rel.)}
& \textbf{Memory (rel.)} \\
\midrule
STGNN            & 1.00$\times$ & 1.00$\times$ & 1.00$\times$ \\
Ghost-Only       & 1.01$\times$ & 1.01$\times$ & 1.01$\times$ \\
Ours-Explicit    & 1.04$\times$ & 1.03$\times$ & 1.02$\times$ \\
Ours-Convexified & 1.08$\times$ & 1.08$\times$ & 1.05$\times$ \\
\bottomrule
\end{tabular}
\vspace{-5pt}
\end{table}

\noindent\textbf{Computational cost.}
As shown in Table~\ref{tab:runtime}, the proposed components introduce
limited overhead.
Ghost fusion operates only around open-boundary nodes, and the local
boundary refiner has a closed-form update.
The explicit global refiner adds one structured post-update, while the
convexified variant further uses two fixed PCG steps.
In practice, the explicit variant is more efficient, while the
convexified variant is more robust for long rollouts and complex meshes.

\medskip
\noindent\textbf{RQ4. Controlled validation of ghost proxy.}
Since ghost nodes are intended to represent unobserved external forcing, they do not have direct ground-truth supervision in real hydrologic systems.
Therefore, our claim is not that they exactly recover the true external process; instead, we aim to use them to learn a surrogate representation of the missing boundary influence that supports stable forecasting.

To validate this interpretation more directly, we conduct a controlled synthetic experiment with known time-varying upstream boundary forcing.
Specifically, we generate data from a linearized 1D shallow-water system on a directed chain graph with 10 real nodes, and impose a known upstream boundary input whose signal consists of a low-frequency sinusoidal background together with pulse components.
We compare our boundary-aware \textit{Ghost} design against a \textit{Virtual Node (VN)} baseline, which connects one global virtual node to all real nodes without explicit boundary structure~\cite{pham2017graph}.

\begin{table}[h]
\centering
\caption{Controlled ghost-proxy validation under known upstream forcing (\(T{=}6\)).}
\label{tab:ghost_synth_compare}
\small
\setlength{\tabcolsep}{8pt}
\renewcommand{\arraystretch}{1.1}
\begin{tabular}{lcc}
\toprule
\textbf{Method} & \textbf{Ghost RMSE} & \textbf{Pearson Corr.} \\
\midrule
\textit{VN} & 0.1018 & 0.206 \\
\textit{Ghost} & \textbf{0.0705} & \textbf{0.961} \\
\bottomrule
\end{tabular}
\vspace{-4pt}
\end{table}

As shown in Table~\ref{tab:ghost_synth_compare}, the proposed \textit{Ghost} substantially outperforms \textit{Virtual Node (VN)} in alignment with the true boundary ghost signal.
It reduces Ghost RMSE from 0.1018 to 0.0705, a relative reduction of about 30.7\%, while increasing Pearson Corr. from 0.206 to 0.961.
Here, Pearson Corr. measures the agreement in temporal variation between the predicted ghost sequence and the true boundary ghost sequence.
These results do not establish strict physical identifiability, but they provide direct and falsifiable evidence that the learned ghost is more than a generic smoothing variable.
In a controlled system with known boundary forcing, it can recover a surrogate strongly aligned with the true boundary influence.

\medskip
\noindent\textbf{RQ5. Extreme/high-flow regimes.}
Aggregate RMSE over the full rollout may obscure events that matter most in hydrologic forecasting, such as severe flooding, storm surges, or rapidly changing high-flow periods.
To address this concern, we evaluate the model under extreme/high-flow regimes using both standard RMSE and importance-weighted RMSE (IW-RMSE).
This metric assigns larger weights to nodes and periods with stronger temporal variation and higher accumulated flow.

For each real node \(i\), let \(r_i(t)\) denote the relative historical flow over the input window.
We define
{\setlength{\abovedisplayskip}{5pt}
\setlength{\belowdisplayskip}{5pt}
\[
d_i=
\frac{1}{T_{\mathrm{in}}}
\sum_{t=1}^{T_{\mathrm{in}}}
|\partial_t r_i(t)|,
\qquad
a_i=\int r_i(t)\,dt.
\]

Then IW-RMSE is computed as
\[
\mathrm{IW\text{-}RMSE}
=
\left[
\frac{1}{N_{\mathrm{real}}T_{\mathrm{out}}}
\sum_{i=1}^{N_{\mathrm{real}}}
\sum_{\tau=1}^{T_{\mathrm{out}}}
d_i^2a_i
(\hat y_{i,\tau}-y_{i,\tau})^2
\right]^{1/2}.
\]
}
The weight \(d_i^2a_i\) emphasizes nodes with stronger temporal variation and higher accumulated flow, which are particularly relevant to flood forecasting.
In implementation, \(d_i^2a_i\) is clipped to \([0.1,10]\) for numerical stability.

\begin{table}[h]
\centering
\captionsetup{skip=2pt}
\caption{Extreme/high-flow evaluation on River System under autoregressive rollout (\(T{=}6\)).}
\label{tab:results_t6_rmse_iw}
\small
\setlength{\tabcolsep}{5.5pt}
\renewcommand{\arraystretch}{1.05}
\begin{tabular}{lcccc}
\toprule
\textbf{Method} & \textbf{RMSE} & \textbf{Imp.} & \textbf{IW} & \textbf{Imp.} \\
\midrule
STGNN & 0.3499 & -- & 0.4423 & -- \\
Ghost-Only & 0.2777 & 20.63\% & 0.2422 & 45.24\% \\
Ours-Conv. & \textbf{0.2311} & \textbf{33.95\%} & \textbf{0.1770} & \textbf{59.98\%} \\
\bottomrule
\end{tabular}
\vspace{-4pt}
\end{table}

As shown in Table~\ref{tab:results_t6_rmse_iw}, compared with the STGNN backbone, Ghost-Only improves RMSE and IW-RMSE by 20.63\% and 45.24\%, respectively.
Ours-Convexified further improves them by 33.95\% and 59.98\%.
The larger gain on IW-RMSE than on standard RMSE indicates that the proposed framework is not only more stable on average, but also more effective in high-impact and dynamically active regimes.
These results also suggest that, although simplified, the physics-guided refiners still provide useful stabilizing structure in the tested extreme/high-flow regime.

\medskip
\noindent\textbf{RQ6. Sensitivity analysis.}
We finally examine whether the Convexified Refiner depends on physical initialization or proxy-network design.
We perturb key learnable physical scales, \(\lambda_{\mathrm{reg}}\), \(\Delta t\), and \(\hat g\), by \(\pm20\%\) at initialization, changing one parameter per run.
We also vary the Proxy MLP width and depth as an additional robustness check.

\begin{table}[t]
\centering
\captionsetup{skip=2pt}
\caption{Sensitivity of the Convexified Refiner.}
\label{tab:conv_sensitivity}
\small
\setlength{\tabcolsep}{4.6pt}
\renewcommand{\arraystretch}{1.02}

\textit{(a) Physical initialization.}
\vspace{2pt}

\begin{tabular}{lcccc}
\toprule
\textbf{Change} & \(\lambda_{\rm reg}\) & \(\Delta t\) & \(\hat g\) & \textbf{RMSE / \(\%\Delta\)} \\
\midrule
Default & 1.0 & 1.0 & 1.0 & 0.2311 / -- \\
\(\lambda_{\rm reg}\) +20\% & 1.2 & 1.0 & 1.0 & 0.2324 / +0.56\% \\
\(\lambda_{\rm reg}\) -20\% & 0.8 & 1.0 & 1.0 & 0.2322 / +0.47\% \\
\(\Delta t\) +20\% & 1.0 & 1.2 & 1.0 & 0.2339 / +1.21\% \\
\(\Delta t\) -20\% & 1.0 & 0.8 & 1.0 & 0.2309 / -0.09\% \\
\(\hat g\) +20\% & 1.0 & 1.0 & 1.2 & 0.2364 / +2.29\% \\
\(\hat g\) -20\% & 1.0 & 1.0 & 0.8 & 0.2348 / +1.60\% \\
\bottomrule
\end{tabular}

\vspace{5pt}
\textit{(b) Proxy MLP architecture.}
\vspace{2pt}

\begin{tabular}{cccc}
\toprule
\textbf{Width} & \textbf{Depth} & \textbf{RMSE} & \(\%\Delta\) \\
\midrule
128 & 3 & 0.2311 & -- \\
256 & 2 & 0.2332 & +0.91\% \\
384 & 2 & 0.2344 & +1.43\% \\
256 & 3 & 0.2306 & -0.22\% \\
384 & 3 & 0.2341 & +1.30\% \\
\bottomrule
\end{tabular}
\vspace{-10pt}
\end{table}

As shown in Table~\ref{tab:conv_sensitivity}(a), the final RMSE varies by at most 2.29\%, indicating that the Convexified Refiner remains robust under deliberately biased physical initializations.
The sensitivity to \(\lambda_{\mathrm{reg}}\) is below 0.6\%, suggesting that the convexified formulation does not rely on a narrowly tuned regularization strength.
Table~\ref{tab:conv_sensitivity}(b) shows that the RMSE variation remains within 1.43\% when changing the Proxy MLP width and depth, further suggesting that the framework is insensitive to moderate architectural changes.
Together, these results provide evidence that the performance gain of \myalg\ is not caused by fragile initialization or a specific proxy-network capacity.

\section{Related Work}
\label{sec:related_work}

\noindent\textbf{\emph{Hydrologic and flood forecasting on meshes/graphs.}}
Graph- and mesh-based surrogates are increasingly used for hydrologic forecasting to represent river topology, irregular discretizations, and time-varying open boundaries.
General architectures such as MeshGraphNets~\cite{Pfaff2021MeshGraphNets}, DCRNN~\cite{li2018dcrnn}, and Graph WaveNet~\cite{wu2019graphwavenet} provide common spatiotemporal backbones, while recent flood models adapt graph learning to hydraulic dynamics and long autoregressive rollouts.
FloodGNN-GRU~\cite{Kazadi2024FloodGNNGRU} combines message passing with recurrent temporal modeling, mSWE-GNN~\cite{bentivoglio2025boundary} targets shallow-water rollout under time-varying/open boundaries, and HydroGAT~\cite{Sarkar2025HydroGAT} scales forecasting to heterogeneous basin-wide river--land graphs.

\noindent\textbf{\emph{Open boundaries in numerical models and learned surrogates.}}
Classical ocean/coastal models handle open boundaries using radiation or characteristic conditions, relaxation/nudging, and sponge layers to stabilize long integrations~\cite{Marchesiello2001OBC,LavelleThacker2008Sponge,BlayoDebreu2005OBC,Marsaleix2006OBC,zhang2016schism}.
Learned surrogates often assume boundary signals are available at rollout time, e.g., ghost-cell-style padding in mSWE-GNN~\cite{bentivoglio2025boundary} and boundary padding in PhyMPGN~\cite{Zeng2025PhyMPGN}.
When external forcings or observations are available, they can also be injected through data fusion~\cite{Sun2022HESS}, while operator-learning methods condition on boundary or discretization information, as in CP-DANO~\cite{Xu2021CPDANO} and MoE-BC~\cite{Deighan2025MoEBC}; related physics-constrained graph learning has also been explored for water systems~\cite{Mu2025STGPINN}.
In contrast, we consider \emph{open-boundary uncertainty}, where external drivers are unavailable at inference time.
Unlike the global virtual node in~\cite{pham2017graph}, which connects bidirectionally to all nodes to form a graph-level representation, our ghost proxies are boundary-specific and inferred from local boundary–interior dynamics to approximate unresolved external forcing.

\noindent\textbf{\emph{Physics refiners for stable long rollouts.}}
Autoregressive forecasts can drift because of compounding model error, imperfect boundary specification, and numerical instability.
Recent correction methods include diffusion-based PDE-Refiner~\cite{lippe2023pde}, training-free PhysicsCorrect~\cite{Huang2025PhysicsCorrect}, and future-aware objectives for long-range simulation~\cite{luo2023fair}.
Related ideas arise in data assimilation, such as EnKF~\cite{Evensen2003EnKF} and hydrologic observation fusion~\cite{Sun2022HESS}, while DUALFloodGNN~\cite{Acosta2025DUALFloodGNN} imposes global/local physical consistency for multi-step flood prediction.
Our physics refiners build on this correction paradigm but specifically regularize propagation under unresolved open-boundary forcing.

\section{Conclusion}

We proposed a new \myalg\ framework for open-boundary spatiotemporal graph forecasting that addresses missing boundary forcing and suppresses error amplification during long-horizon autoregressive rollout.
Our framework learns ghost-node proxies for unobserved boundary forcing, stabilizes local proxy--boundary consistency with a boundary refiner, and regularizes boundary-driven propagation with a global physics refiner.
Because simplified PDE priors may not fully capture real-world hydrologic complexity, we further introduce a convexified semi-implicit refiner.
It performs stable corrective updates from the boundary-refined prediction, rather than solving the full PDE to convergence.
Extensive experiments on two real open-boundary hydrologic systems demonstrate consistent improvements in rollout accuracy and stability. Future work will investigate abrupt external forcing, together with uncertainty-aware forecasting under such events.

\section{acknowledgment}

This work was supported in part by the National Science Foundation (NSF) under Grant Nos.\,IIS-2544353 (CAREER), IIS-2505719, IIS-2236579, IIS-2302786, IIS-2441449, IOS-2430224, and IOS-2446522, the U.S. Environmental Protection Agency (EPA) under Grant No.\,EPA-R3-CBP-21-08, the Science Center for Marine Fisheries, and the Commonwealth Cyber Initiative (CCI). 

\bibliographystyle{IEEEtran}
\bibliography{references}

@article{zhang2016schism,
  author  = {Zhang, Yinglong J. and Ye, Fei and Stanev, Emil V. and Grashorn, Sebastian},
  title   = {Seamless cross-scale modeling with {SCHISM}},
  journal = {Ocean Modelling},
  volume  = {102},
  pages   = {64--81},
  year    = {2016}
}

@article{klingler2021lamah,
  author  = {Klingler, Christoph and Schulz, Karsten and Herrnegger, Mathew},
  title   = {LamaH-CE: {LArge-SaMple DAta} for hydrology and environmental sciences for Central Europe},
  journal = {Earth System Science Data},
  volume  = {13},
  pages   = {4529--4565},
  year    = {2021}
}

@article{bentivoglio2025boundary,
  author  = {Bentivoglio, Roberto and Isufi, Elvin and Jonkman, Sebastiaan Nicolas and Taormina, Riccardo},
  title   = {Multi-scale hydraulic graph neural networks for flood modelling},
  journal = {Natural Hazards and Earth System Sciences},
  volume  = {25},
  number  = {1},
  pages   = {335--351},
  year    = {2025}
}

@article{mccabe2023stability,
  author  = {McCabe, Michael and Harrington, Peter and Subramanian, Shashank and Brown, Jed},
  title   = {Towards Stability of Autoregressive Neural Operators},
  journal = {Transactions on Machine Learning Research},
  year    = {2023}
}

@inproceedings{li2018dcrnn,
  author    = {Li, Yaguang and Yu, Rose and Shahabi, Cyrus and Liu, Yan},
  title     = {Diffusion Convolutional Recurrent Neural Network: Data-Driven Traffic Forecasting},
  booktitle = {ICLR},
  year      = {2018}
}

@inproceedings{wu2019graphwavenet,
  author    = {Wu, Zonghan and Pan, Shirui and Long, Guodong and Jiang, Jing and Zhang, Chengqi},
  title     = {Graph {WaveNet} for Deep Spatial-Temporal Graph Modeling},
  booktitle = {IJCAI},
  pages     = {1907--1913},
  year      = {2019}
}

@article{BergerLeVeque98,
  author  = {Berger, Marsha J. and LeVeque, Randall J.},
  title   = {Adaptive Mesh Refinement Using Wave-Propagation Algorithms for Hyperbolic Systems},
  journal = {SIAM Journal on Numerical Analysis},
  volume  = {35},
  number  = {6},
  pages   = {2298--2316},
  year    = {1998}
}

@inproceedings{lippe2023pde,
  author    = {Lippe, Phillip and Veeling, Bastiaan S. and Perdikaris, Paris and Turner, Richard E. and Brandstetter, Johannes},
  title     = {{PDE}-Refiner: Achieving Accurate Long Rollouts with Neural {PDE} Solvers},
  booktitle = {NeurIPS},
  pages     = {67398--67433},
  year      = {2023}
}

@inproceedings{Huang2025PhysicsCorrect,
  author    = {Huang, Xinquan and Perdikaris, Paris},
  title     = {{PhysicsCorrect}: A Training-Free Approach for Stable Neural {PDE} Simulations},
  booktitle = {AAAI},
  pages     = {22057--22065},
  year      = {2026}
}

@inproceedings{jiang2025topology,
  author    = {Jiang, Haoyang and Wang, Jindong and Zhu, Xingquan and He, Yi},
  title     = {Topology-aware Neural Flux Prediction Guided by Physics},
  booktitle = {ICML},
  pages     = {27670--27684},
  year      = {2025}
}

@inproceedings{kirschstein2024merit,
  author    = {Kirschstein, Nikolas and Sun, Yixuan},
  title     = {The Merit of River Network Topology for Neural Flood Forecasting},
  booktitle = {ICML},
  pages     = {24713--24725},
  year      = {2024}
}

@article{cui2024total,
  author  = {Cui, Linlin and Ye, Fei and Zhang, Y. Joseph and Yu, Haocheng and Wang, Zhengui and Moghimi, Saeed and Seroka, Gregory and Riley, Jack and Pe'eri, Shachak and Mani, Soroosh and Myers, Edward and Park, Kyungmin and Tang, Liujuan and Yang, Zizang and Wang, Yan-Ming},
  title   = {Total Water Level Prediction at Continental Scale: Coastal Ocean},
  journal = {Ocean Modelling},
  volume  = {192},
  pages   = {102451},
  year    = {2024}
}

@article{tseng2003ghost,
  author  = {Tseng, Yu-Heng and Ferziger, Joel H.},
  title   = {A ghost-cell immersed boundary method for flow in complex geometry},
  journal = {Journal of Computational Physics},
  volume  = {192},
  number  = {2},
  pages   = {593--623},
  year    = {2003}
}

@book{vreugdenhil1994numerical,
  author    = {Vreugdenhil, Cornelis Boudewijn},
  title     = {Numerical Methods for Shallow-Water Flow},
  publisher = {Kluwer Academic Publishers},
  year      = {1994}
}

@book{leveque2002finite,
  author    = {LeVeque, Randall J.},
  title     = {Finite Volume Methods for Hyperbolic Problems},
  publisher = {Cambridge University Press},
  year      = {2002}
}

@article{ascher1997implicit,
  author  = {Ascher, Uri M. and Ruuth, Steven J. and Spiteri, Raymond J.},
  title   = {Implicit-explicit Runge-Kutta methods for time-dependent partial differential equations},
  journal = {Applied Numerical Mathematics},
  volume  = {25},
  number  = {2--3},
  pages   = {151--167},
  year    = {1997}
}

@article{Kazadi2024FloodGNNGRU,
  author  = {Kazadi, Arnold and Doss-Gollin, James and Sebastian, Antonia and Silva, Arlei},
  title   = {{FloodGNN-GRU}: a spatio-temporal graph neural network for flood prediction},
  journal = {Environmental Data Science},
  volume  = {3},
  pages   = {e21},
  year    = {2024}
}

@inproceedings{Poli2020GraphNeuralODE,
  author    = {Poli, Michael and Massaroli, Stefano and Park, Junyoung and Yamashita, Atsushi and Asama, Hajime and Park, Jinkyoo},
  title     = {Graph Neural Ordinary Differential Equations},
  booktitle = {AAAIW},
  year      = {2020}
}

@inproceedings{Li2024SSMTemporalGraphs,
  author    = {Li, Jintang and Wu, Ruofan and Jin, Xinzhou and Ma, Boqun and Chen, Liang and Zheng, Zibin},
  title     = {State Space Models on Temporal Graphs: A First-Principles Study},
  booktitle = {NeurIPS},
  pages     = {127030--127058},
  year      = {2024}
}

@inproceedings{Gravina2024LongRangeCTDG,
  author    = {Gravina, Alessio and Lovisotto, Giulio and Gallicchio, Claudio and Bacciu, Davide and Grohnfeldt, Claas},
  title     = {Long Range Propagation on Continuous-Time Dynamic Graphs},
  booktitle = {ICML},
  pages     = {16206--16225},
  year      = {2024}
}

@inproceedings{Kong2024STPGNN,
  author    = {Kong, Weiyang and Guo, Ziyu and Liu, Yubao},
  title     = {Spatio-Temporal Pivotal Graph Neural Networks for Traffic Flow Forecasting},
  booktitle = {AAAI},
  pages     = {8627--8635},
  year      = {2024}
}

@inproceedings{luo2023fair,
  author    = {Luo, Xiao and Luo, Junyu and Jiang, Huiyu and Zhou, Hang and Xiao, Zhiping and Ju, Wei and Yang, Carl and Zhang, Ming and Sun, Yizhou},
  title     = {Future Matters for Present: Towards Effective Physical Simulation over Meshes},
  booktitle = {KDD},
  pages     = {906--917},
  year      = {2025}
}

@inproceedings{pham2017graph,
  author    = {Pham, Trang and Tran, Truyen and Dam, Hoa Khanh and Venkatesh, Svetha},
  title     = {Graph Classification via Deep Learning with Virtual Nodes},
  booktitle = {IJCAIW},
  year      = {2017}
}

@inproceedings{Pfaff2021MeshGraphNets,
  author    = {Pfaff, Tobias and Fortunato, Meire and Sanchez-Gonzalez, Alvaro and Battaglia, Peter W.},
  title     = {Learning Mesh-Based Simulation with Graph Networks},
  booktitle = {ICLR},
  year      = {2021}
}

@inproceedings{Sarkar2025HydroGAT,
  author    = {Sarkar, Aishwarya and Hakimi, Autrin and Chen, Xiaoqiong and Huang, Hai and Lu, Chaoqun and Demir, Ibrahim and Jannesari, Ali},
  title     = {{HydroGAT}: Distributed Heterogeneous Graph Attention Transformer for Spatiotemporal Flood Prediction},
  booktitle = {SIGSPATIAL},
  pages     = {1019--1030},
  year      = {2025}
}

@article{Marchesiello2001OBC,
  author  = {Marchesiello, Patrick and McWilliams, James C. and Shchepetkin, Alexander F.},
  title   = {Open Boundary Conditions for Long-Term Integration of Regional Oceanic Models},
  journal = {Ocean Modelling},
  volume  = {3},
  number  = {1--2},
  pages   = {1--20},
  year    = {2001}
}

@article{LavelleThacker2008Sponge,
  author  = {Lavelle, James W. and Thacker, William C.},
  title   = {A Pretty Good Sponge: Dealing with Open Boundaries in Limited-Area Ocean Models},
  journal = {Ocean Modelling},
  volume  = {20},
  number  = {3},
  pages   = {270--292},
  year    = {2008}
}

@article{BlayoDebreu2005OBC,
  author  = {Blayo, Eric and Debreu, Laurent},
  title   = {Revisiting Open Boundary Conditions from the Point of View of Characteristic Variables},
  journal = {Ocean Modelling},
  volume  = {9},
  number  = {3},
  pages   = {231--252},
  year    = {2005}
}

@article{Marsaleix2006OBC,
  author  = {Marsaleix, Patrick and Auclair, Francis and Estournel, Claude},
  title   = {Considerations on Open Boundary Conditions for Regional and Coastal Ocean Models},
  journal = {Journal of Atmospheric and Oceanic Technology},
  volume  = {23},
  number  = {11},
  pages   = {1604--1613},
  year    = {2006}
}

@inproceedings{Zeng2025PhyMPGN,
  author    = {Zeng, Bocheng and Wang, Qi and Yan, Mengtao and Liu, Yang and Chengze, Ruizhi and Zhang, Yi and Liu, Hongsheng and Wang, Zidong and Sun, Hao},
  title     = {{PhyMPGN}: Physics-Encoded Message Passing Graph Network for Spatiotemporal {PDE} Systems},
  booktitle = {ICLR},
  year      = {2025}
}

@article{Sun2022HESS,
  author  = {Sun, Alexander Y. and Jiang, Peishi and Yang, Zong-Liang and Xie, Yangxinyu and Chen, Xingyuan},
  title   = {A Graph Neural Network ({GNN}) Approach to Basin-Scale River Network Learning: The Role of Physics-Based Connectivity and Data Fusion},
  journal = {Hydrology and Earth System Sciences},
  volume  = {26},
  number  = {19},
  pages   = {5163--5184},
  year    = {2022}
}

@inproceedings{Xu2021CPDANO,
  author    = {Xu, Jiayang and Pradhan, Aniruddhe and Duraisamy, Karthikeyan},
  title     = {Conditionally Parameterized, Discretization-Aware Neural Networks for Mesh-Based Modeling of Physical Systems},
  booktitle = {NeurIPS},
  pages     = {1634--1645},
  year      = {2021}
}

@article{Deighan2025MoEBC,
  author  = {Deighan, Dwyer and Actor, Jonas A. and Patel, Ravi G.},
  title   = {Mixture of Neural Operator Experts for Learning Boundary Conditions and Model Selection},
  journal = {arXiv preprint arXiv:2502.04562},
  year    = {2025}
}

@article{Mu2025STGPINN,
  author  = {Mu, Tianwei and Duan, Feiyu and Ning, Baokuan and Zhou, Bo and Liu, Junyu and Huang, Manhong},
  title   = {{ST-GPINN}: a spatio-temporal graph physics-informed neural network for enhanced water quality prediction in water distribution systems},
  journal = {npj Clean Water},
  volume  = {8},
  pages   = {74},
  year    = {2025}
}

@article{Evensen2003EnKF,
  author  = {Evensen, Geir},
  title   = {The Ensemble Kalman Filter: Theoretical Formulation and Practical Implementation},
  journal = {Ocean Dynamics},
  volume  = {53},
  number  = {4},
  pages   = {343--367},
  year    = {2003}
}

@inproceedings{Acosta2025DUALFloodGNN,
  author    = {Acosta, Carlo Malapad and Herath, Herath Mudiyanselage Viraj Vidura and Lim, Jia Yu and Saha, Abhishek and Rasnayaka, Sanka and Marshall, Lucy},
  title     = {{DUALFloodGNN}: Physics-Informed Graph Neural Network for Operational Flood Modeling},
  booktitle = {IJCAI},
  year      = {2026}
}

\end{document}